\documentclass[journal]{IEEEtran}
\usepackage{graphicx}
\usepackage{amstext}
\usepackage{algorithm}
\usepackage{algorithmic}
\usepackage{mathrsfs}
\usepackage{amssymb}
\usepackage{amsmath}
\usepackage{bm}
\allowdisplaybreaks[4]
\usepackage{epstopdf}
\usepackage{multicol}
\usepackage{stfloats}
\usepackage{url}
\usepackage{color}
\usepackage{enumerate}
\usepackage{breqn}
\usepackage{bbm}
\usepackage{cite}
\usepackage[font=small]{caption}
\usepackage{subcaption}
\usepackage{enumitem}
\usepackage{array}
\usepackage{comment}
\usepackage{epsfig}
\usepackage{pifont}

\usepackage{tabularx}
\usepackage{booktabs}

\ifCLASSINFOpdf
\else
\fi
\begin{document}
	
	\title{Smart IoT-Based Wearable Device for Detection and Monitoring of Common Cow Diseases Using a Novel Machine Learning Technique}

	\author{Rupsa Rani Mishra, D. Chandrasekhar Rao, Ajaya Kumar Tripathy$^{*}$
		
		\thanks{Rupsa Rani Mishra, D. Chandrasekhar Rao are with the Department of CSE, CUPGS, Biju Patnaik University, India.(Emails: rupsarani1@gmail.com; cupgs.dcrao@bput.ac.in,).}
		\thanks{Ajaya Kumar Tripathy is with the Department of Computer Science Gangadhar Meher University, Odisha, India. (Emails: ajayatripathy1@gmail.com).}

		\thanks{Corresponding author:Ajaya Kumar Tripathy}
	}

	
	\markboth{IEEE Journal Name }%
	{Shell \MakeLowercase{\textit{et al.}}: Bare Demo of IEEEtran.cls for IEEE Journals}
	
	\maketitle

	\begin{abstract}
		Manual observation and monitoring of individual cows for disease detection present significant challenges in large-scale farming operations, as the process is labor-intensive, time-consuming, and prone to reduced accuracy. The reliance on human observation often leads to delays in identifying symptoms, as the sheer number of animals can hinder timely attention to each cow. Consequently, the accuracy and precision of disease detection are significantly compromised, potentially affecting animal health and overall farm productivity.
		Furthermore, organizing and managing human resources for the manual observation and monitoring of cow health is a complex and economically demanding task. It necessitates the involvement of skilled personnel, thereby contributing to elevated farm maintenance costs and operational inefficiencies.
		Therefore, the development of an automated, low-cost, and reliable smart system is essential to address these challenges effectively. Although several studies have been conducted in this domain, very few have simultaneously considered the detection of multiple common diseases with high prediction accuracy. However, advancements in Internet of Things (IoT), Machine Learning (ML), and Cyber-Physical Systems have enabled the automation of cow health monitoring with enhanced accuracy and reduced operational costs. This study proposes an IoT-enabled Cyber-Physical System framework designed to monitor the daily activities and health status of cow. 
		A novel ML algorithm is proposed for the diagnosis of common cow diseases using collected physiological and behavioral data. The algorithm is designed to predict multiple diseases by analyzing a comprehensive set of recorded physiological and behavioral features, enabling accurate and efficient health assessment.
		Comparative analysis with existing methods demonstrates that the proposed algorithm achieves superior predictive performance, highlighting its effectiveness and reliability.
	\end{abstract}

	
	\IEEEpeerreviewmaketitle

	\section{Introduction}
	\IEEEPARstart{D}{airy} cow farming constitutes a crucial segment of the agricultural industry, playing a significant role in economic development. It provides farmers with a stable source of income through the production of milk, dairy products, and manure—essential commodities that are in high demand across both domestic and international markets. This demand contributes positively to economic growth and sustainability \cite{george1996dairying}.
	As herd sizes grow, manual observation of individual cow becomes increasingly impractical, leading to delays in detecting behavioral changes, early signs of illness, and potential disease outbreaks. Traditional methods, reliant on human labor, are not only resource-intensive but also prone to errors that can compromise animal welfare and farm productivity. 
	Automated technologies, including sensor-based health monitoring and predictive analytics, are therefore critical for improving the precision, efficiency, and responsiveness of farm management. Integrating such systems enables proactive health interventions, reduces economic losses associated with disease, and enhances the overall sustainability of large-scale dairy operations.

\begin{figure}[tbh]
	\centering
	\includegraphics[width=0.49\textwidth]{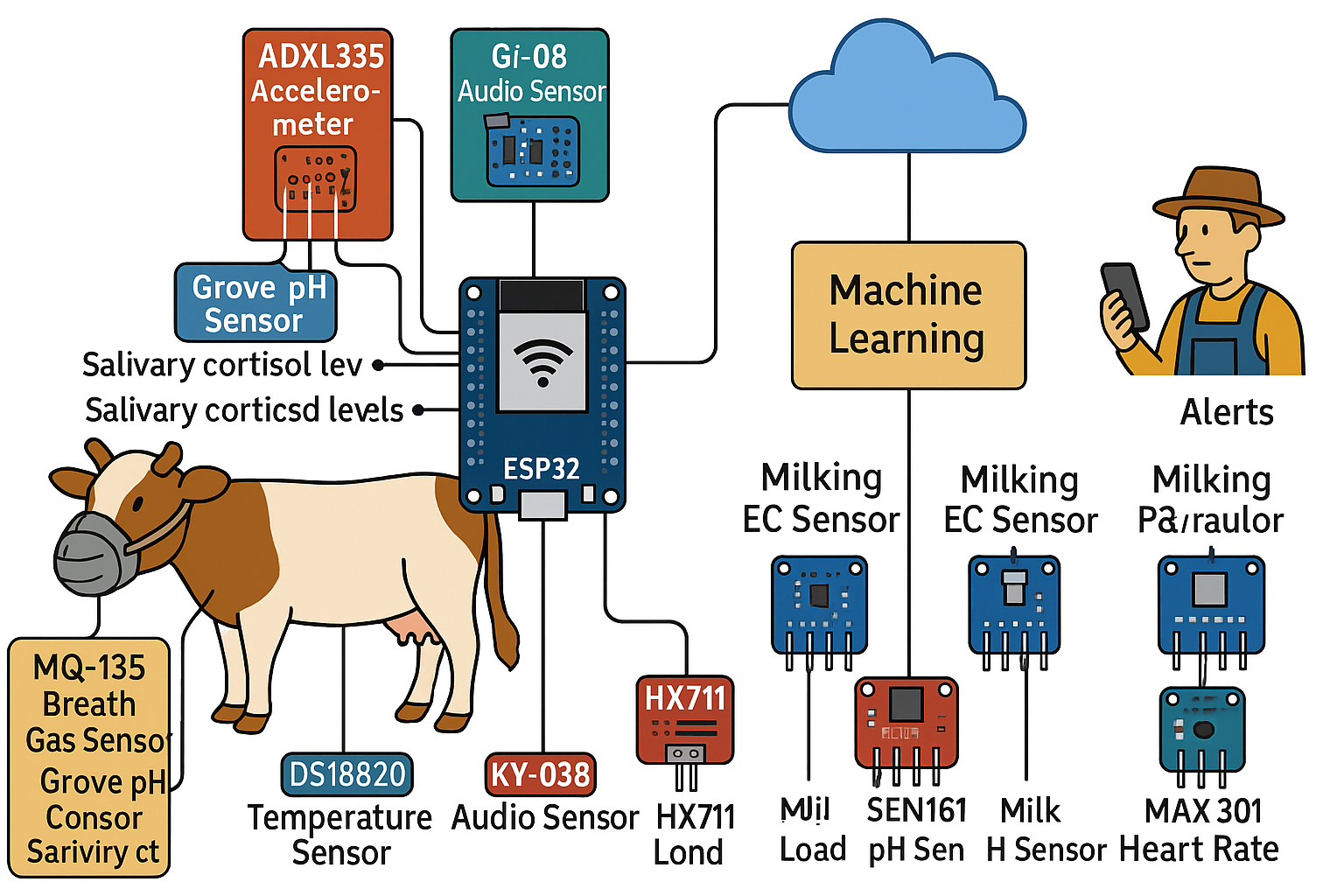}
	\caption{Smart Dairy Cow Farm Cyber-Physical System.}
	\label{fig:arch}
\end{figure}

	The challenges associated with large-scale dairy cow farming, particularly in health monitoring and disease prediction, necessitate the adoption of advanced technological solutions. Internet of Things (IoT) devices, such as wearable sensors and automated monitoring systems, enable real-time collection of physiological and behavioral data from individual animals. When integrated with Machine Learning (ML) algorithms, this data can be analyzed to detect anomalies, predict potential health issues, and optimize decision-making processes with greater accuracy than traditional methods. Furthermore, the development of a comprehensive cyber-physical system (CPS) is vital for managing and controlling farm operations remotely. By linking physical processes with IoT-based computational networks, a CPS facilitates real-time data collection, feature extraction, health monitoring, and both automated and remote access and control, thereby enhancing the precision, efficiency, and resilience of large-scale dairy farming systems.  
	Automated cow health monitoring and disease prediction systems are particularly beneficial for dairy farms located in remote areas, where access to veterinary services is limited. By enabling early detection of common diseases, these systems support timely intervention and treatment, thereby improving animal health outcomes and reducing potential economic losses.

	Recent advancements in Internet of Things (IoT) technologies have profoundly impacted the evolution of smart ecosystems and agricultural practices. These innovations have facilitated the development of interconnected, cost-effective solutions that not only improve operational efficiency but also contribute to enhanced productivity and overall quality of life \cite{senoo2024iot,tripathy2021mygreen,mohanty2021internet,sethuraman2022idrone,liu2020industry,chatterjee2021livecare,sharma2020machine,friha2021internet,bushby2024early,setser2024individuality}. IoT-enabled systems have shown promise in automating livestock health monitoring by utilizing sensors such as heart rate monitors, accelerometers, and wearable devices to detect physiological and behavioral changes associated with disease \cite{alonso2020intelligent,sharma2018cattle,awasthi2016non,vyas2019fmd,chandra2022cattlecare}.
	
	Moreover, the integration of ML with IoT has emerged as a powerful approach in precision livestock farming. Notable studies have demonstrated success in tasks such as body condition scoring \cite{yukun2019automatic}, diagnosis of respiratory and metabolic diseases \cite{casella2023machine,feighelstein2024ai}, and early detection of Digital Dermatitis using behavioral and thermal imaging data \cite{magana2023machine}. Although a wide range of IoT applications have been effectively implemented in agriculture and dairy cow farming \cite{arif2024resource,dutta2025internet,tangorra2024internet,bordignon2025smart}, the advancement of integrated IoT-based CPS specifically designed for real-time, multi-disease prediction in dairy cows remains relatively underexplored. To address this gap, the present study proposes a novel CPS framework that utilizes multimodal sensor data in conjunction with ML algorithms to enable individualized health assessment, early disease detection, and continuous health monitoring.
	
	The proposed cyber-physical system integrates a network of sensors strategically deployed on the cow's neck belt and at key external locations such as the walking path, milking parlor, and solid mouth mask to continuously capture individual cows' physiological and behavioral parameters. These sensors transmit the collected signals to a cloud server via wireless communication channels. The cloud infrastructure performs signal preprocessing and feature extraction to derive relevant health indicators from the raw sensor data.
	
	The extracted features are utilized by a ML-based predictive model specifically developed for the detection of common bovine diseases, including Bloat, Bovine Respiratory Disease (BRD), Displaced Abomasum, Foot and Mouth Disease (FMD), Hardware Disease, Johne's Disease, Ketosis, Lameness, Mastitis, Milk Fever, Tuberculosis, and Acidosis. The predictive model is trained on a labeled dataset comprising historical sensor data annotated with disease outcomes. Once the model achieves satisfactory classification accuracy, it is deployed for real-time prediction.
	
	Upon receiving real-time feature inputs from the sensor network, the model assesses the cow’s health status and identifies the presence of any disease. Farmers can access comprehensive health diagnostics and status reports remotely via a user-friendly, web-based interface. The overall semantics of the automated dairy cow health monitoring CPS is illustrated in Fig. \ref{fig:arch}.

	The remainder of the article is structured as follows. Section~\ref{sec:CPS} details the proposed CPS architecture. Section~\ref{sec:HPOSVM} outlines the proposed ML model for common cow disease detection. Section~\ref{sec:ExperimentalEva} presents the experimental setup and results analysis. Finally, Section~\ref{sec:con} concludes the paper and discusses potential directions for future research.
	
\section{A Cyber-Physical System for Personalized Health Monitoring in Dairy Cows}
\label{sec:CPS}
The primary objectives of the proposed CPS are to automate and optimize the process of individualized cow observation and health monitoring. The system is specifically designed to reduce the need for continuous human supervision by incorporating intelligent automation, thereby minimizing human involvement in routine health assessments. A critical focus is on ensuring on-time or early detection of probable diseases through advanced data analytics and ML techniques. Furthermore, the CPS is developed to facilitate seamless remote access and control of the health monitoring system, enabling stakeholders to observe, assess, and respond to the health conditions of cows efficiently, regardless of physical location.

The automation of the health monitoring process without the continuous involvement of human experts can be achieved by employing sensor networks to capture relevant behavioral and physiological data. These data are transmitted to a cloud server via wireless communication protocols, where they are subsequently analyzed using ML algorithms to identify patterns indicative of potential diseases. Upon detecting any anomalies associated with health risks, the system generates real-time reports. These reports are accessible remotely by stakeholders, including farmers, veterinarians, and field experts, enabling them to make timely and informed decisions for prompt intervention.

\begin{figure}[tbh]
	\centering
	\includegraphics[width=0.48\textwidth, trim=0cm 1cm 0cm 5cm,clip]{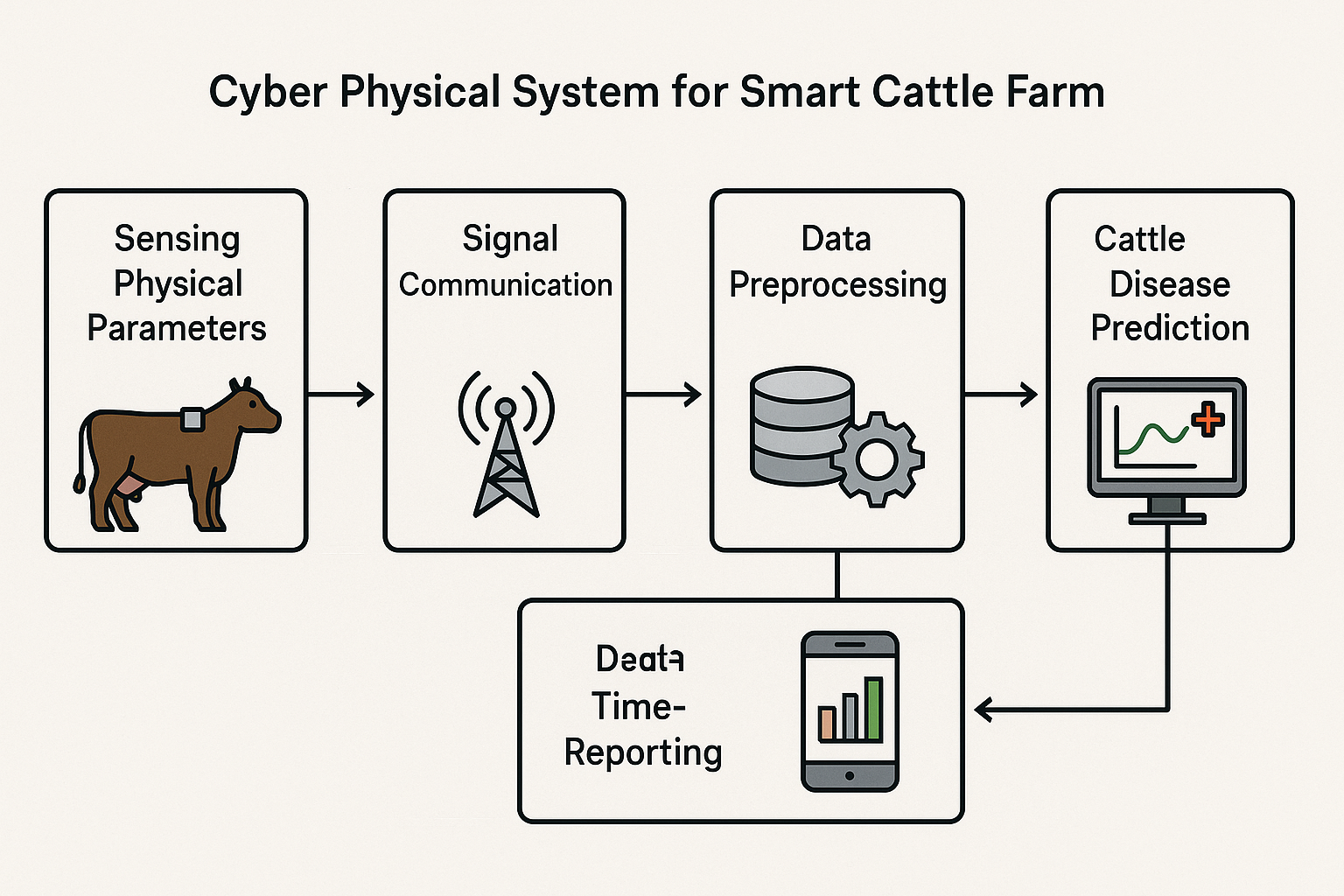}
	\caption{Overview of the working model of the proposed CPS.}
	\label{fig:cps}
\end{figure}

A central element of the proposed CPS is a machine-learning-based disease prediction module, which utilizes a novel multi-class classification algorithm for accurate identification of various health conditions. This framework enables the generation of real-time alerts through integrated mobile and web applications, thereby supporting the early diagnosis and prompt treatment of affected animals. The overall working model of the proposed CPS for smart cow farming is illustrated in Fig. \ref{fig:cps}. The system comprises five primary components: (i) sensing of physical parameters, (ii) wireless signal communication, (iii) data preprocessing and feature extraction, (iv) disease prediction using ML models, and (v) real-time reporting and alert dissemination to relevant stakeholders.

\begin{figure}[tbh]
	\centering
	\includegraphics[width=0.49\textwidth]{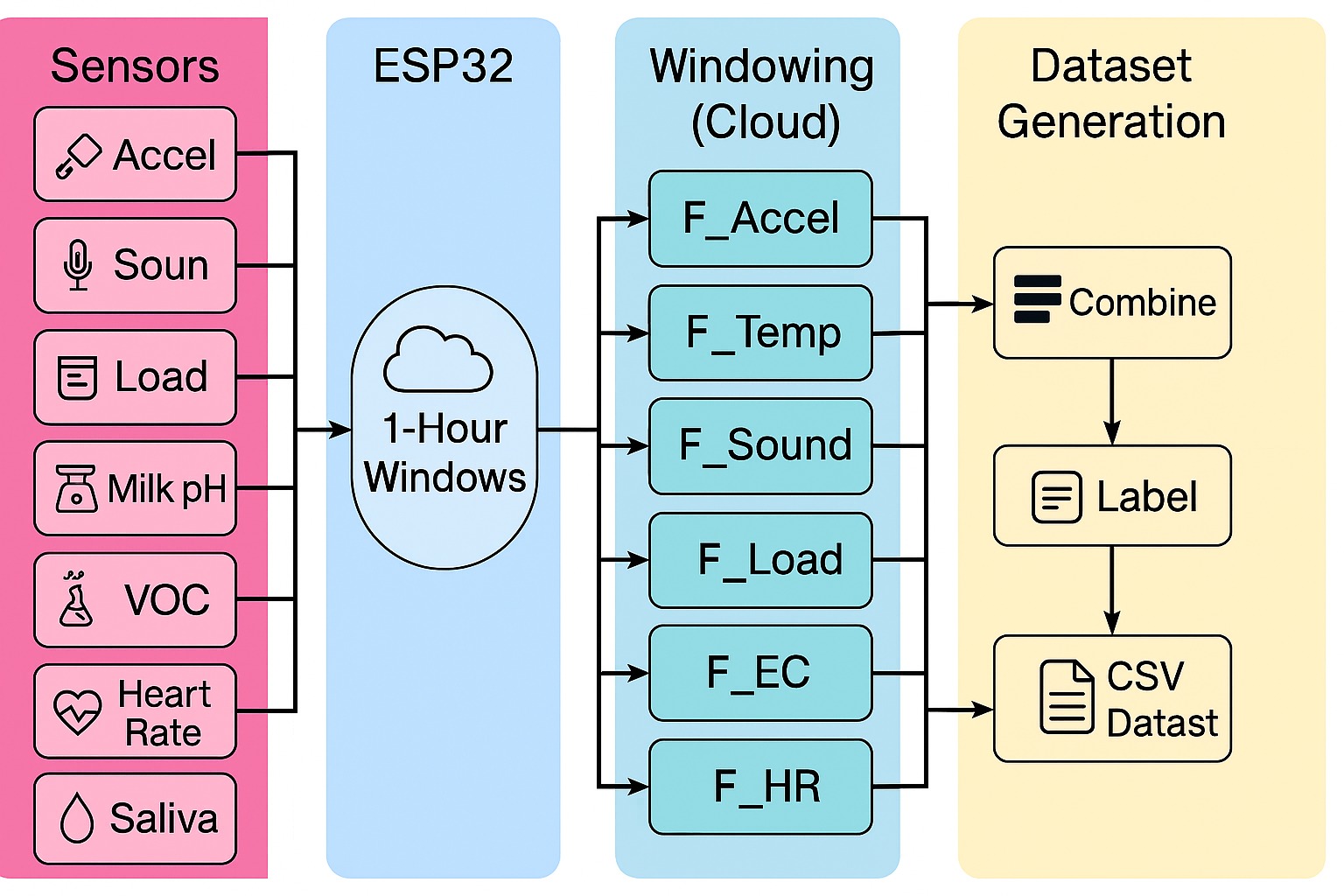}
	\caption{Architecture of the proposed cyber-physical system (CPS) for automated cow health monitoring.}
	\label{fig:arch1}
\end{figure}

\subsection{Wearable Sensing Model for Recording Physical Parameters}
The proposed CPS is designed to facilitate real-time monitoring and early detection of twelve prevalent cow diseases. These include: Bloat, BRD, Displaced Abomasum, FMD, Hardware Disease, Johne's Disease, Ketosis, Lameness, Mastitis, Milk Fever, Tuberculosis, and Acidosis.
The symptoms associated with these diseases are primarily detected through an array of sensors integrated within the CPS. These includes Accelerometer,
Temperature Sensor, Audio Sensor, Load Sensor, 
Milk Electrical Conductivity (EC) Sensor, Milk pH Sensor, Breath Gas Sensor, 
 Heart Rate Sensor, Saliva pH Sensor, and Salivary Cortisol Sensor.
 The sensor network enables continuous physiological and behavioral monitoring, facilitating timely identification of disease indicators.  The architecture of the proposed CPS is presented Fig \ref{fig:arch1}. The list of sensors, along with their specifications and the features extracted from each, is provided in TABLE~\ref{tab:sensors}.

\begin{table*}[ht]
	
	\caption{Specifications and Extracted Features of Cow Health Monitoring Sensors}
	\centering
	\begin{tabularx}{\textwidth}{l l X}
		\toprule
		\textbf{Sensor Type} & \textbf{Sensors and Specifications} & \textbf{Extracted Features} \\
		\midrule
		Accelerometer & ADXL335 (3-axis, $\pm$2 to $\pm$16g range, I2C interface) & 
		Lying Time (hrs/day), Steps/day, Standing$\rightarrow$Lying Transitions/day, Activity Index (0–100) \\
		
		Temperature Sensor & DS18B20 (Digital, -55\textdegree C to +125\textdegree C, $\pm$0.5\textdegree C accuracy) & 
		Average, Max, and Min Body Temperature (\textdegree C), Fever Duration $>$39.5\textdegree C (hrs/day) \\
		
		Audio Sensor & KY-038 (Microphone sensor module, analog/digital output) & 
		Cough Count/day, Moo Count/day, Avg Moo Duration (sec), Moo Pitch (Hz), Cough Amplitude (dB) \\
		
		Load Sensor & HX711 + Load Cell (up to 200kg, analog to digital output) & 
		Leg Load Imbalance (\% diff), Load Imbalance Duration (min/day) \\
		
		Milk EC Sensor & Milking EC Sensor (1–15 mS/cm, IP-rated) & 
		Day Avg EC (mS/cm), EC Asymmetry across udder quarters, Max EC from any quarter \\
		
		Milk pH Sensor & SEN0161 & 
		Day Avg pH, Low pH Events/day (pH $<$ 6.4), High pH Events/day (pH $>$ 6.8) \\
		
		Breath Gas Sensor & MQ-135 (Detects NH\textsubscript{3}, CO\textsubscript{2}, CH\textsubscript{4}, analog output) & 
		NH\textsubscript{3} (ppm), CH\textsubscript{4} (ppm), H\textsubscript{2}S (ppm), VOC Index (0–500) \\
		
		Heart Rate Sensor & MAX30100 (Pulse oximeter and heart rate sensor, digital output) & 
		Heart Rate (bpm), HRV (ms, SDNN approx.), SpO\textsubscript{2} (\%) \\
		
		Saliva pH Sensor & Grove pH Sensor & 
		Saliva pH Range \\
		
		Salivary Cortisol Sensor & ELISA-based cortisol assay (e.g., Salimetrics) & 
		Salivary Cortisol Level (ng/mL) \\
		\bottomrule
	\end{tabularx}
	
	\label{tab:sensors}
	\end{table*}

\subsubsection{Hardware Architecture and Sensor Integration}
The proposed IoT-based cow health monitoring system is designed around the ESP32 microcontroller, which provides a compact and power-efficient platform with integrated Wi-Fi for real-time data transmission and remote monitoring capabilities. Several sensors are deployed on the neck belt, leg belt, and in environmental zones such as the mouth mask, milking parlour, and walking ramp to capture critical physiological and behavioral parameters of the dairy cow.
An analog-output accelerometer (ADX335) is strategically mounted at both the neck and leg of the animal to monitor activity patterns and postural shifts indicative of locomotion disorders or lameness. These sensors interface directly with the analog-to-digital converter (ADC) channels of the ESP32. Core body temperature is measured using a DS18B20 digital temperature sensor affixed to the neck, which communicates via the 1-Wire protocol, utilizing a dedicated GPIO pin with a 4.7\,k$\Omega$ pull-up resistor for signal stability.
A KY-038 audio sensor, also placed at the neck, captures acoustic signatures such as coughing or abnormal vocalizations. Its analog signal is processed via the ESP32’s ADC input. Body weight is continuously monitored through a load cell integrated with an HX711 amplifier module, installed on a walking ramp. The HX711 interfaces with the ESP32 using two digital GPIOs (DT and SCK), enabling precise weight tracking as the animal moves.
In the milking parlour, milk EC is measured using a dedicated analog EC sensor, and milk pH is monitored via the SEN0161 pH sensor. Both sensors connect to separate ADC inputs and provide key indicators for early mastitis detection. Cardiovascular health is assessed using a MAX30100 sensor, which measures heart rate and peripheral blood oxygen saturation (\textit{SpO\textsubscript{2}}). The sensor communicates with the ESP32 through the shared I$^2$C bus (SDA and SCL lines), supported by 4.7\,k$\Omega$ pull-up resistors to ensure robust data transfer.
To evaluate respiratory and oral health, a set of sensors is embedded within a custom-designed mouth mask. A gas sensor (MQ-135) detects volatile compounds such as ammonia and methane in exhaled breath, while a Grove analog pH sensor measures salivary pH as an indicator of feeding behavior and ruminal function. Additionally, a non-invasive salivary cortisol sensor, based on ELISA (Enzyme-Linked Immunosorbent Assay) principles, is employed to assess stress hormone levels. The ELISA sensor output is either digitized via an external ADC or processed through a signal conditioning circuit before interfacing with the ESP32.
The system is powered by a 3.7\,V lithium-polymer battery, regulated through a TP4056 charging module for safe recharging operations. A 5\,V boost converter supplies sensors requiring higher voltage, while all 3.3\,V components are powered via the ESP32’s onboard regulator. This integrated hardware architecture enables comprehensive, real-time, multi-parameter health monitoring, supporting proactive animal welfare management and precision livestock farming.

\subsection{Signal Communication}

Signal communication from the deployed sensors to the cloud infrastructure is facilitated through a centralized microcontroller unit, such as the ESP32, which serves as the primary node for data acquisition, processing, and transmission. Each sensor, depending on its output modality, is interfaced via appropriate communication protocols: analog sensors (e.g., KY-038 audio sensor, MQ-135 breath gas sensor, and Grove pH sensor) utilize the ESP32’s built-in ADC channels, while digital sensors (e.g., DS18B20, MAX30100, and HX711) communicate via standardized protocols such as I\textsuperscript{2}C, SPI, and 1-Wire. The ELISA-based salivary cortisol sensor, though semi-automated, is designed to transmit results through a serial or digital interface post-analysis. Raw signals from all sensors undergo local pre-processing, including noise filtering and data normalization, to ensure consistency and reliability. The processed data are structured into lightweight JSON payloads and transmitted to a cloud-based server using secure communication protocols such as MQTT or HTTPS, enabled by the ESP32’s integrated Wi-Fi module. TLS/SSL encryption ensures data confidentiality and integrity during transmission. To enhance robustness, the system incorporates asynchronous communication, acknowledgment protocols, and retry mechanisms to mitigate data loss under unstable network conditions. For remote farm environments lacking reliable Wi-Fi coverage, alternative communication modules such as GSM (SIM800L) and LoRa are integrated, supporting energy-efficient, scalable data transmission to the cloud for further analytics and decision support.

\begin{figure}[bth]
	\centering
	\includegraphics[width=0.49\textwidth]{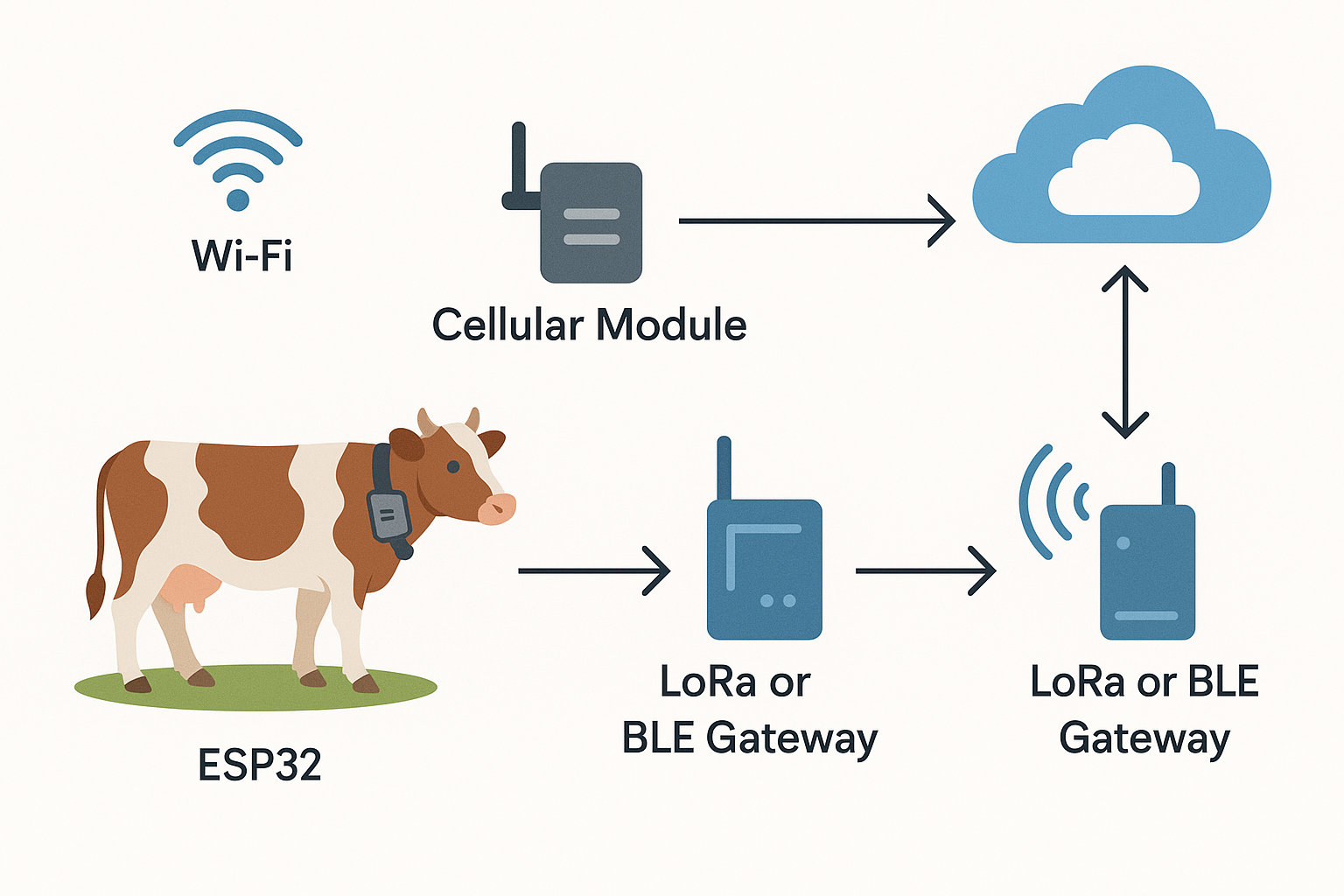}
	\caption{ESP32 data transmission via Wi-Fi, Cellular, or LoRa/BLE Gateway to the cloud.}
	\label{fig:communication}
\end{figure}

\subsection{Preprocessing and Feature Generation}
\label{sec:features}
Sensor data are collected and transmitted to a centralized cloud server for processing. Individual sensor readings, as well as combinations of multiple sensor inputs, are utilized to extract physiologically and behaviorally relevant features associated with common disease symptoms. These features constitute the input set for disease prediction models.

\subsection{Cow Disease Prediction}
The extracted features from each individual cow are input into a pre-trained ML model to predict the presence or absence of disease, thereby determining whether the cow is healthy or affected. To enable early and reliable disease detection, a novel classification algorithm specifically designed for cow disease prediction is proposed. A comprehensive description of the algorithm's design is provided in Section~\ref{sec:HPOSVM}.

\subsection{Real-Time Reporting}
At regular time intervals, the system collects relevant features for each individual cow and performs disease prediction in real time. The health status of each cow is continuously updated based on the prediction results. In the event of a status change, the system automatically notifies the appropriate personnel—such as farmers, veterinary doctors, or caretakers—via a real-time reporting portal, auto-generated emails, and text messages to facilitate timely preventive action. All stakeholders have access to the online reporting platform, where they can monitor the health status, view historical data, and collaborate to decide on appropriate preventive or corrective measures.

\section{HPOSVM: A Hyperparameter-Optimized Support Vector Machine for Early Detection of Common Cow Diseases}
\label{sec:HPOSVM}
This section presents a novel ML approach, the Hyperparameter-Optimized Support Vector Machine (HPOSVM), designed for the early detection of common cow diseases. The proposed method enhances the traditional Support Vector Machine (SVM) by integrating a genetic algorithm-based hyperparameter optimization framework. This optimization mechanism systematically fine-tunes the SVM parameters to improve classification accuracy and model robustness, thereby enabling more reliable and timely disease prediction.

The physiologically and behaviorally relevant features associated with common diseases are used as input to the disease prediction algorithm. A pre-collected dataset comprising these features along with corresponding disease labels is utilized to train the model, enabling it to learn the underlying patterns associated with various health conditions.

The collection of feature-label pairs is defined as \( D = \{ (F_i, l_i) \}_{i=1}^{N} \), where \( F_i \in \mathbb{R}^{P} \) represents a feature vector with \( P \) predictors, and \( l_i \) is the corresponding disease label associated with \( F_i \), referred to as the class label.

  For the time being, we consider the problem of distinguishing between healthy and unhealthy cows, which is formulated as a binary classification task. Accordingly, the class label satisfies \( l_i \in \{ +1, -1 \} \), where \( +1 \) indicates a healthy status and \( -1 \) indicates an unhealthy condition.
Assuming that the two classes are linearly separable, a separating hyperplane can be formulated as 
\[
\sum Z^i F^i + a = 0
\]
or equivalently,
\[
Z^T F + a = 0,
\]
where \( Z \) represents the adjustable weight matrix, \( a \in \mathbb{R} \)  is the bias term, and \( F \) is the feature vector.

\begin{equation*}
	\therefore Z^{T}F_{i}+a \geq 0 ~if~ l_{i}=+1;
\end{equation*}

\begin{equation*}
	Z^{T}F_{i}+a \leq 0 ~if~ l_{i}=-1
\end{equation*}
Although infinitely many hyperplanes can separate the two classes, the objective is to identify the hyperplane that maximizes the margin from the nearest data points. This hyperplane is referred to as the optimal hyperplane for class separation.

The objective is to determine the optimal \( Z \) and \( a \) such that  
\[
l_i (Z^{T} F_i + a) \geq 1 \quad \text{for all} \quad i = 1,2,\ldots,N,
\]
while simultaneously minimizing the objective function
\[
\Phi(Z) = \frac{1}{2} \lVert Z \rVert^{2}.
\]
Here, \( \lVert Z \rVert \) denotes the Euclidean norm of the weight vector \( Z \).

To solve this optimization problem, we employ the method of Lagrangian multipliers. The corresponding Lagrangian formulation is given by

\begin{equation}
	f(Z, a, \beta) = -\sum_{i=1}^{N} \beta_{i} \left[ l_{i} (Z^T F_{i} + a) - 1 \right] + \frac{1}{2} \lVert Z \rVert^{2},
	\label{eq:2}
\end{equation}

where \( \beta_{i} \geq 0 \) are the Lagrangian multipliers associated with each constraint.

Differentiating the Lagrangian function with respect to \( Z \) and setting the result equal to zero yields

\begin{equation}
	Z = \sum_{i=1}^{N} \beta_{i} l_{i} F_{i},
	\label{eq:3}
\end{equation}

where \( l_i \) denotes the class label corresponding to \( F_i \).  

Similarly, differentiating the Lagrangian with respect to \( a \) and equating to zero gives the constraint

\begin{equation}
	\sum_{i=1}^{N} \beta_{i} l_{i} = 0.
	\label{eq:4}
\end{equation}

Therefore, equation~\ref{eq:2} transforms into the following form:

\begin{equation}
	f(\beta)=\sum_{i=1}^{N}\beta_{i}-\frac{1}{2}\sum_{i=1}^{N}\sum_{j=1}^{N}\beta_{i}\beta_{j}l_{i}l_{j}F_{i}^{T}F_{j}
	\label{eq:5}
\end{equation}
where $\beta_{i} \geq 0$.

Given the training set \( \left\{ (F_{i}, l_{i}) \right\}_{i=1}^{N} \), the task is to find the set of Lagrange multipliers \( \left\{ \beta_{i} \right\}_{i=1}^{N} \) that maximize the dual objective function \( f(\beta) \), while satisfying the conditions
\[
\sum_{i=1}^{N} \beta_{i} l_{i} = 0, \quad \text{and} \quad \beta_{i} \geq 0 \quad \text{for} \quad all \quad samples.
\]

The solution of the above constrained optimization problem results in the determination of the optimal separating hyperplane.

To effectively handle nonlinearity present in the dataset, each input sample is projected into a higher-dimensional feature space through an appropriate nonlinear mapping function \( \phi \), where \( \phi(F) \in \mathbb{R}^{P'} \) and \( P' > P \).

In the transformed higher-dimensional feature space \( \mathbb{R}^{P'} \), the data points become linearly separable. Consequently, in equation~\ref{eq:5}, the inner product \( (F_{i}^T  F_{j}) \) is replaced by \( (\phi(F_{i})^T \phi(F_{j})) \). However, explicitly computing the nonlinear mapping \( \phi(F) \) is often computationally prohibitive. 
 
However, instead of explicitly computing the inner product \( \phi(F_{i}) \cdot \phi(F_{j}) \) in the high-dimensional feature space, a kernel function \( g(F_{i}, F_{j}) \) can be employed. By substituting \( (F_{i} \cdot F_{j}) \) with \( g(F_{i}, F_{j}) \) in the optimization formulation, the problem can be efficiently solved without directly computing the nonlinear mapping.

Subject to the constraints \( \sum_{i=1}^{N} \beta_{i} l_{i} = 0 \) and \( 0 \leq \beta_{i} \leq r \), the dual optimization problem can be formulated as:

\[
\arg\max_{\beta} f(\beta) = \arg\max_{\beta} \left( \sum_{i=1}^{N} \beta_{i} - \frac{1}{2} \sum_{i=1}^{N} \sum_{j=1}^{N} \beta_{i} \beta_{j} l_{i} l_{j} g(F_{i}, F_{j}) \right)
\]
\noindent
where \( r \) denotes the regularization parameter that controls the trade-off between maximizing the margin and minimizing classification error.

The discriminant function \( D(F) \) used for classification can be expressed as:

\[
D(F) = \sum_{i=1}^{N} \beta_{i} l_{i} g(F_{i}, F) + a
\]
\noindent
where \( \beta_{i} \) are the Lagrange multipliers, \( l_{i} \) are the corresponding class labels, \( g(F_{i}, F) \) denotes the kernel function evaluating the similarity between the support vector \( F_{i} \) and the input sample \( F \), and \( a \) represents the bias term.

Although this approach effectively manages nonlinearity within the dataset, it presents difficulties in determining optimal hyperparameters, such as kernel parameters and the regularization coefficient. To address this, a Genetic Algorithm (GA)-based optimization strategy is employed for automatic hyperparameter tuning.

The GA addresses the challenge of identifying the optimal hypothesis from a pool of candidate solutions, where the optimal hypothesis is defined as the one maximizing a predefined fitness function. GA follows an iterative process, beginning with an initial population of hypotheses. In each generation, the fitness of each hypothesis is evaluated, and a new population is formed by selecting the fittest individuals and generating offspring through genetic operations. The newly generated hypotheses are propagated to subsequent generations to progressively improve solution quality.

The fitness function quantifies the quality of each hypothesis within a given generation. Genetic operations, including crossover and mutation, are critical in enhancing the diversity and quality of the evolving hypotheses. Selection mechanisms, such as roulette wheel selection or tournament selection, prioritize higher-quality hypotheses for participation in the crossover process. This evolutionary process is repeated iteratively until a predefined termination criterion is satisfied.

Instead of relying on manually predefined kernel parameters for SVM, the GA explores a hypothesis space of candidate solutions. In this study, the Radial Basis Function (RBF) kernel is selected to illustrate the proposed approach. The RBF kernel requires the optimization of two key parameters, \( C \) and \( \gamma \). A GA-based strategy is developed to automatically optimize these parameters for enhanced classification performance. In the proposed framework, each chromosome encodes the parameters \( C \) and \( \gamma \), with binary encoding employed for representation. The classification accuracy serves as the fitness function guiding the evolutionary search.

The chromosome structure is defined as \((b_{C}^{p}, b_{C}^{p-1}, \dots, b_{C}^{1}, b_{\gamma}^{q}, b_{\gamma}^{q-1}, \dots, b_{\gamma}^{1})\), comprising a total of \(p + q\) binary bits, where \(b\) denotes a binary unit. Here, \(p\) and \(q\) represent the number of bits assigned to the parameters \(C\) and \(\gamma\), respectively. These values can be adjusted to control the precision of the parameter representation. The precision of the encoded parameters is determined by the bit-lengths \(p\) and \(q\), along with the predefined minimum and maximum bounds for \(C\) and \(\gamma\). The actual numerical value of each parameter is obtained through the decoding process described in equation \ref{equn:pheno}.

\begin{equation}
	\label{equn:pheno}
	\alpha_{val}=\frac{[\alpha_{max}-\alpha_{min}]}{2^{k}-1}* \sum_{i=1}^{k}b_{\alpha}^{i}*2^{i-1}
\end{equation}

The actual value of the parameter \(\alpha\) encoded in the chromosome is determined by equation \(\ref{equn:pheno}\), where \(\alpha_{val}\) represents the decoded value of \(\alpha\). The maximum and minimum permissible values of \(\alpha\) are denoted by \(\alpha_{max}\) and \(\alpha_{min}\), respectively. The parameter \(k\) refers to the number of bits allocated for encoding \(\alpha\), and \(b_{\alpha}^{i}\) represents the \(i^{th}\) bit in the binary sequence corresponding to \(\alpha\). The sum over the bits \(b_{\alpha}^{i}\) is weighted by the powers of 2, and the resulting value is scaled to the range \([\alpha_{min}, \alpha_{max}]\).
Fitness Function: The fitness function in this study is based on pixel labeling accuracy, with higher accuracy values corresponding to higher fitness scores. Hypotheses with greater fitness are more likely to be selected for the next generation, promoting the evolution of optimal solutions.
Stopping Criteria: The evolutionary process is halted when one of the following conditions is met: (i) accuracy approaches 95\%, (ii) no significant improvement in accuracy occurs over a predefined number of generations, or (iii) the maximum number of iterations (5000) is reached. These criteria ensure the algorithm converges without running indefinitely.
Parameter Settings for GA: The GA is configured with a population size of 200, a crossover rate of 0.75, and a mutation rate of 0.01. A two-point crossover strategy is employed, and parent selection is conducted using the Roulette wheel method. These parameter settings are designed to balance exploration and exploitation within the search space, facilitating the effective optimization of kernel parameters.

\section{Experimental setup and Performance Evaluation}
\label{sec:ExperimentalEva}
To evaluate the effectiveness of the proposed approach, comprehensive experiments were conducted using a real-world dataset collected from five distinct local cow farms between December 2022 and January 2025. The study focused on prevalent bovine diseases, including Bloat, BRD, Displaced Abomasum, FMD, Hardware Disease, Johne's Disease, Ketosis, Lameness, Mastitis, Milk Fever, Tuberculosis, and Acidosis. The dataset encompasses both diseased and healthy cows, with expert-labeled disease diagnoses. Features corresponding to these diseases and healthy cows were specifically recorded to support model training.

\begin{figure}[H]
	\centering
	\begin{subfigure}[b]{0.24\textwidth}
		\centering
		\includegraphics[width=\textwidth]{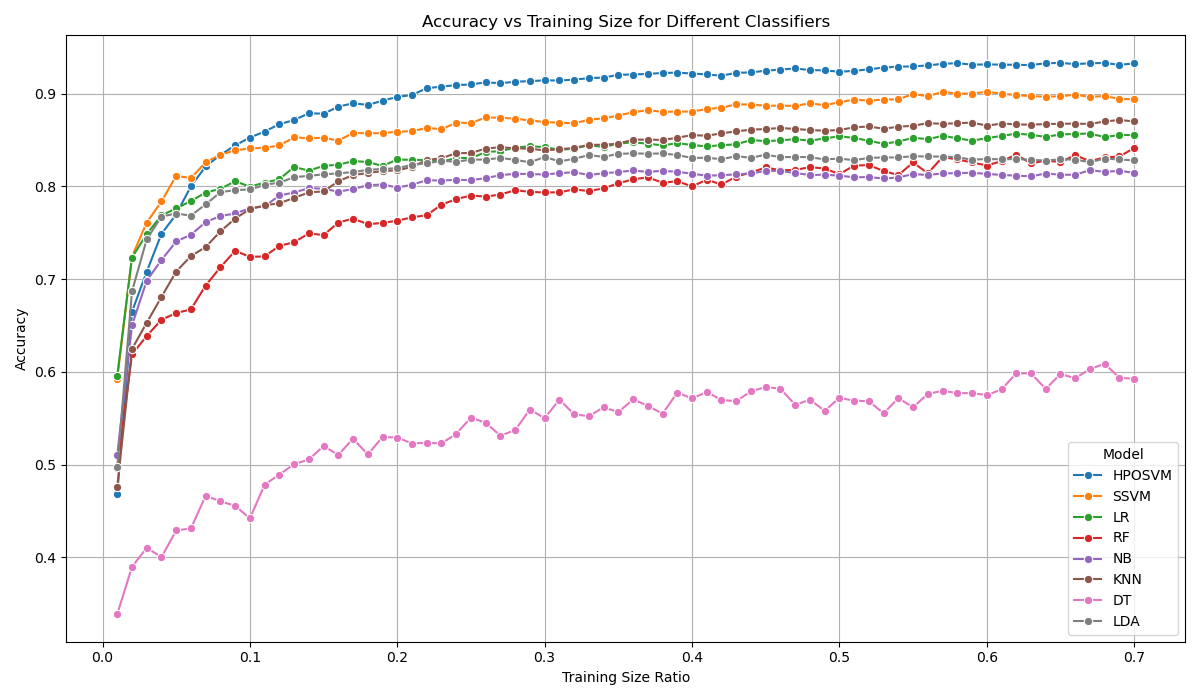}
		\caption{Accuracy}
		\label{fig:accuracy}
	\end{subfigure}
	\hfill
	\begin{subfigure}[b]{0.24\textwidth}
		\centering
		\includegraphics[width=\textwidth]{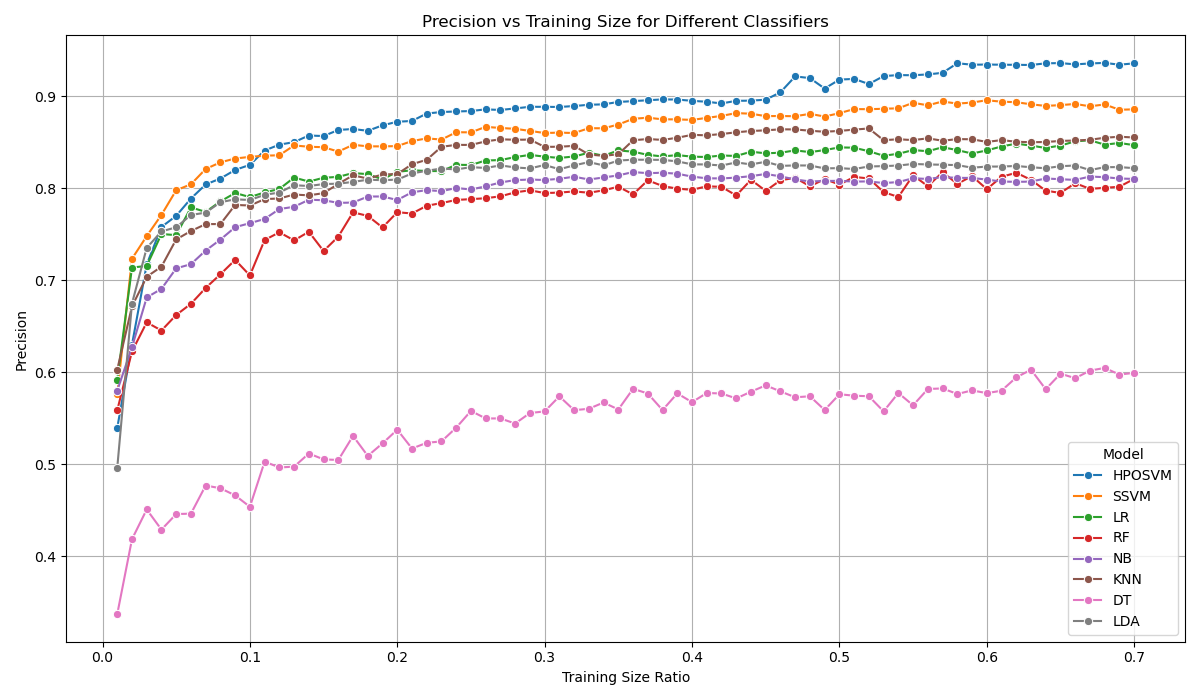}
		\caption{Precision}
		\label{fig:precision}
	\end{subfigure}
	
	\vspace{0.5cm} 
	
	\begin{subfigure}[b]{0.24\textwidth}
		\centering
		\includegraphics[width=\textwidth]{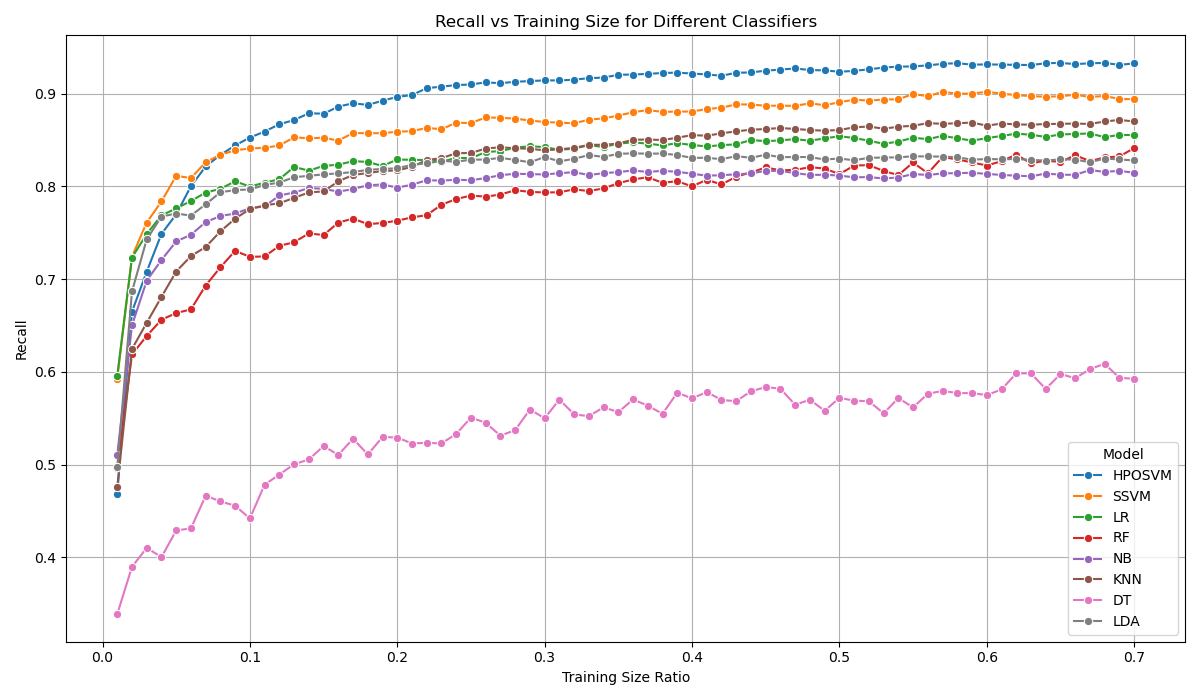}
		\caption{Recall}
		\label{fig:recall}
	\end{subfigure}
	\hfill
	\begin{subfigure}[b]{0.24\textwidth}
		\centering
		\includegraphics[width=\textwidth]{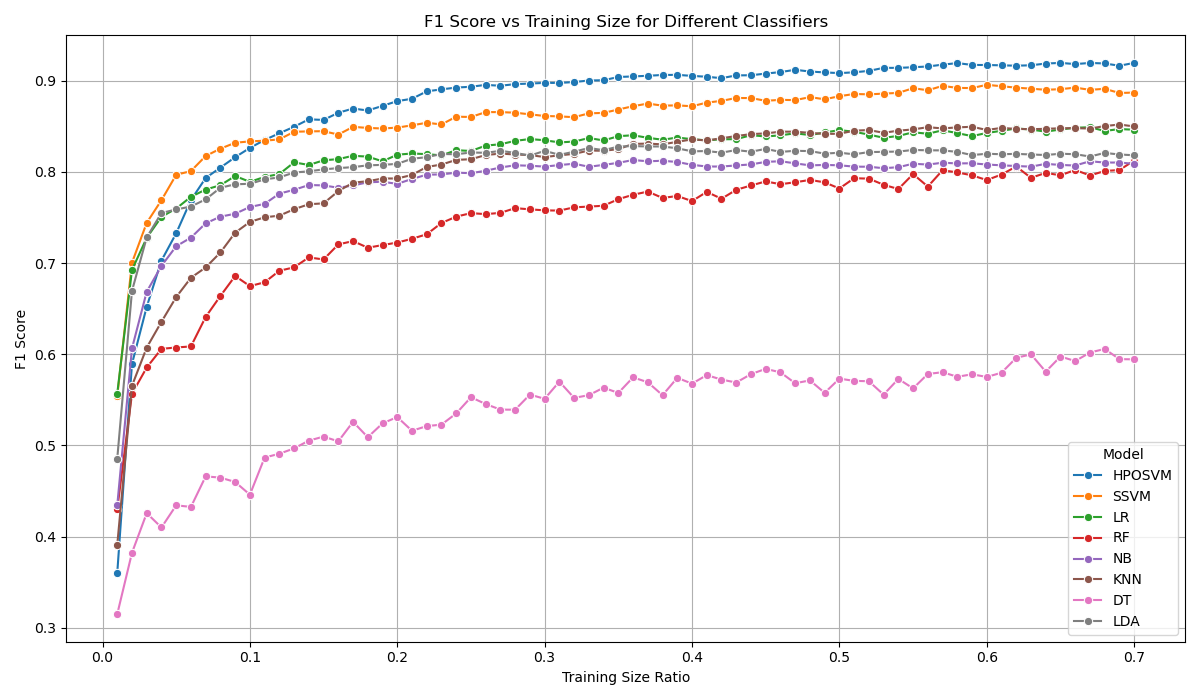}
		\caption{F1 Score}
		\label{fig:f1score}
	\end{subfigure}
	
	\begin{subfigure}[b]{0.24\textwidth}
		\centering
		\includegraphics[width=\textwidth]{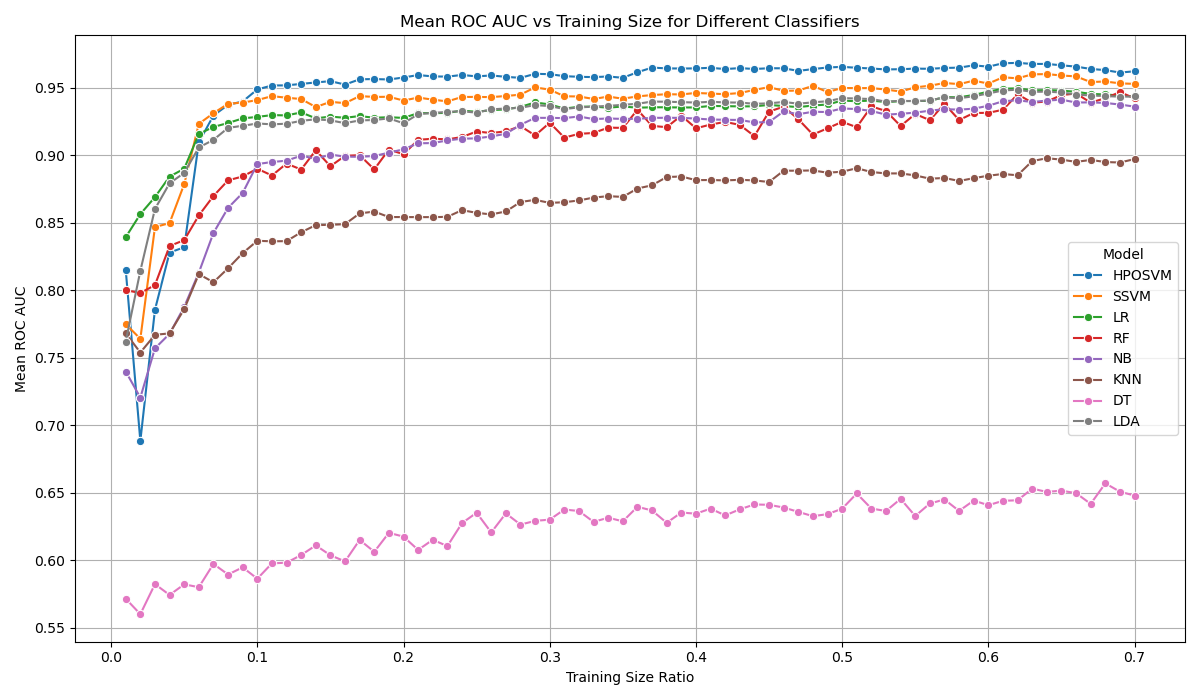}
		\caption{F1 Score}
		\label{fig:ROC-AUC}
	\end{subfigure}
	
	\caption{Performance metrics (Accuracy, Precision, Recall, F1 Score) across training data percentages from 1\% to 70\%.}
	\label{fig:metrics}
\end{figure}

The dataset comprises 30 features, including: (1) Lying Time (hrs/day), (2) Steps/day, (3) Standing→Lying Transitions/day, (4) Activity Index (0–100), (5) Average Temperature (°C), (6) Maximum Temperature (°C), (7) Minimum Temperature (°C), (8) Fever Duration $>$39.5°C (hrs/day), (9) Cough Count/day, (10) Moo Count/day, (11) Average Moo Duration (sec), (12) Moo Pitch (Hz), (13) Cough Amplitude (dB), (14) Leg Load Imbalance (\% difference), (15) Load Imbalance Duration (min/day), (16) Day Average EC (mS/cm), (17) EC Asymmetry (mS/cm) among the four udder quarters, (18) Maximum EC (mS/cm) from any udder quarter during the day, (19) Day Average pH, (20) Low pH Events/day (Count of times pH $<$ 6.4), (21) High pH Events/day (Count of times pH $>$ 6.8), (22) Ammonia (NH$_3$) concentration (ppm), (23) Methane (CH$_4$) concentration (ppm), (24) Hydrogen Sulfide (H$_2$S) concentration (ppm), (25) Volatile Organic Compounds (VOC) Index (0–500), (26) Heart Rate (beats per minute), (27) Heart Rate Variability (HRV) (ms, SDNN approx.), (28) Blood Oxygen Saturation (SpO$_2$) (\%), (29) Saliva pH Range, and (30) Salivary Cortisol Level (ng/mL).

To evaluate the performance of the proposed classifier, the dataset was divided into two subsets: a training set and a testing set. The training data were utilized to train the classification algorithm and construct the predictive model.
After training the classifier, the test dataset is employed to assess its predictive accuracy. The evaluation results are presented using popular classifier accuracy measures like: overall accuracy, precision, recall,  F1-score, and ROC-AUC value are computed. The F1-score is employed as the fitness function within the genetic algorithm to optimize the hyperparameters. 

In addition to the proposed classification model, we conducted experiments using seven widely adopted classification approaches: 
Simple Support Vector Machine (SSVM),
k-Nearest Neighbors (KNN),     
Logistic Regression (LR), 
Linear Discriminant Analysis (LDA), 
Random Forest (RF),
Naive Bayes (NB), and   Decision Tree (DT).

To evaluate the impact of training data volume on model performance, the proportion of training data was systematically varied from 1\% to 70\% in 1\% increments. This approach enabled a granular analysis of how incremental increases in training data affect key metrics. The resulting performance measures—Accuracy, Precision, Recall, and F1 Score—are presented in Fig.~\ref{fig:metrics}, illustrating the correlation between training data size and model efficacy. The analysis reveals that HPOSVM achieves its highest accuracy on test data when trained with 70\% of the dataset, demonstrating superior overall performance.

\begin{table}[htbp]
	\centering
	\caption{Performance comparison of classifiers at 70\% training data}
	\begin{tabular}{p{1.5cm}p{1cm}p{1cm}p{1cm}p{1cm}p{1cm}}
		\hline
		\textbf{Model} & \textbf{Accuracy} & \textbf{Precision} & \textbf{Recall} & \textbf{F1 Score} & \textbf{ROC AUC} \\
		\hline
		HPOSVM              & 0.9327 & 0.9357 & 0.9327 & 0.9196 & 0.9618 \\
		Simple SVM          & 0.8940 & 0.8858 & 0.8940 & 0.8871 & 0.9521 \\
		KNN                 & 0.8700 & 0.8552 & 0.8700 & 0.8499 & 0.8971 \\
		LR & 0.8553 & 0.8469 & 0.8553 & 0.8463 & 0.9434 \\
		LDA                 & 0.8280 & 0.8219 & 0.8280 & 0.8183 & 0.9435 \\
		RF       & 0.8220 & 0.7961 & 0.8220 & 0.7903 & 0.9510 \\
		Naive Bayes         & 0.8147 & 0.8102 & 0.8147 & 0.8089 & 0.9359 \\
		DT       & 0.5960 & 0.5985 & 0.5960 & 0.5960 & 0.6509 \\
		\hline
	\end{tabular}
	\label{tab:performance70}
\end{table}

\begin{figure}[htbp]
	\centering
	\includegraphics[width=\linewidth]{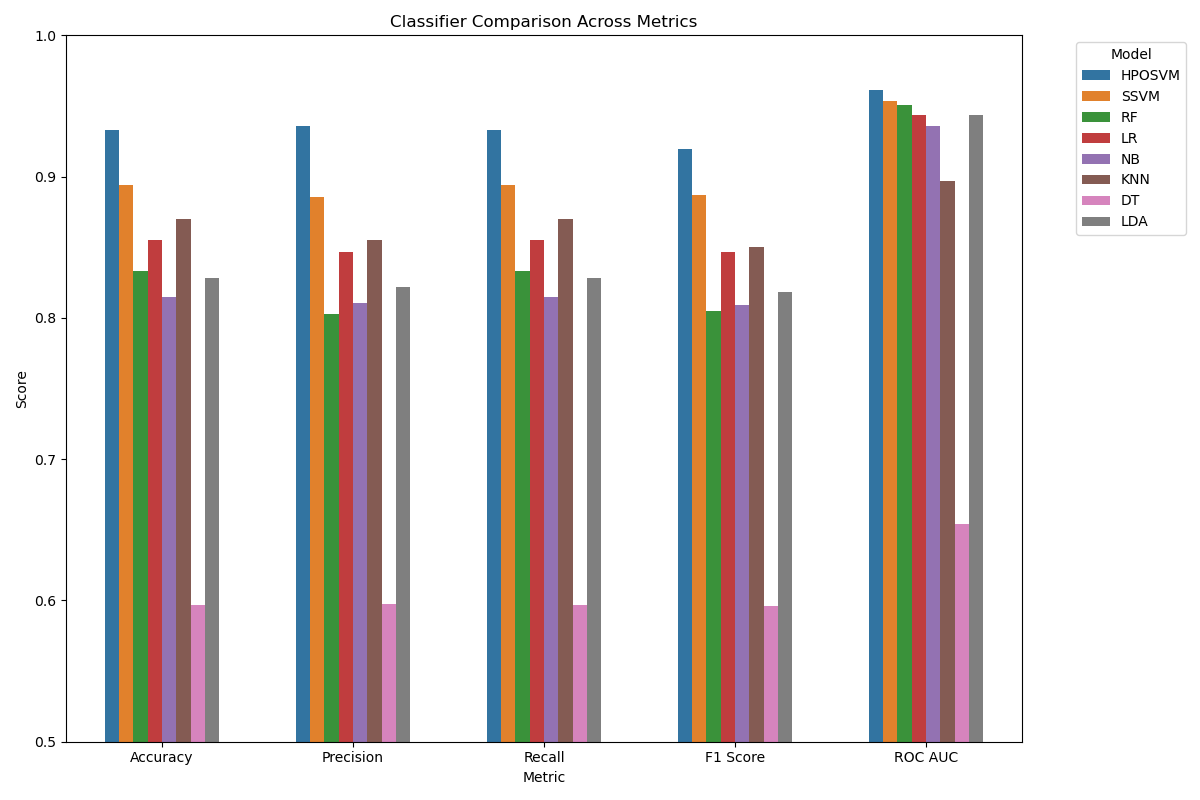}
	\caption{Metric-wise performance comparison of the proposed HPOSVM classifier with baseline models.}
	\label{fig:classifier_comparison}
\end{figure}

The performance metrics on the test dataset, obtained from the model trained with 70\% of the data, are detailed in TABLE~\ref{tab:performance70}.
To facilitate a comparative analysis, Fig.~\ref{fig:classifier_comparison} visually presents the metric-wise performance of various classifiers alongside the proposed classifier. This visualization aids in assessing the relative efficacy of each model across key evaluation metrics.

The ROC AUC metric evaluates the classifier’s ability to distinguish between positive and negative instances by measuring how well it ranks positive cases above negative ones across varying decision thresholds. The ROC curves illustrating the performance of each classifier are presented in the following Fig.~\ref{fig:roc_curves_all} .

\begin{figure}[htbp]
	\centering
	\begin{subfigure}[b]{0.24\textwidth}
		\centering
		\includegraphics[width=\textwidth]{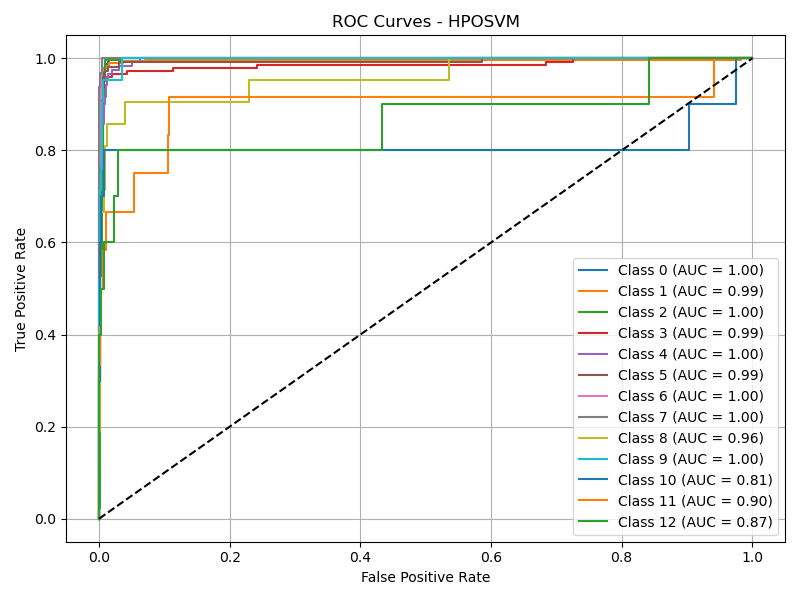}
		\caption{HPOSVM}
	\end{subfigure}
	\hfill
	\begin{subfigure}[b]{0.24\textwidth}
		\centering
		\includegraphics[width=\textwidth]{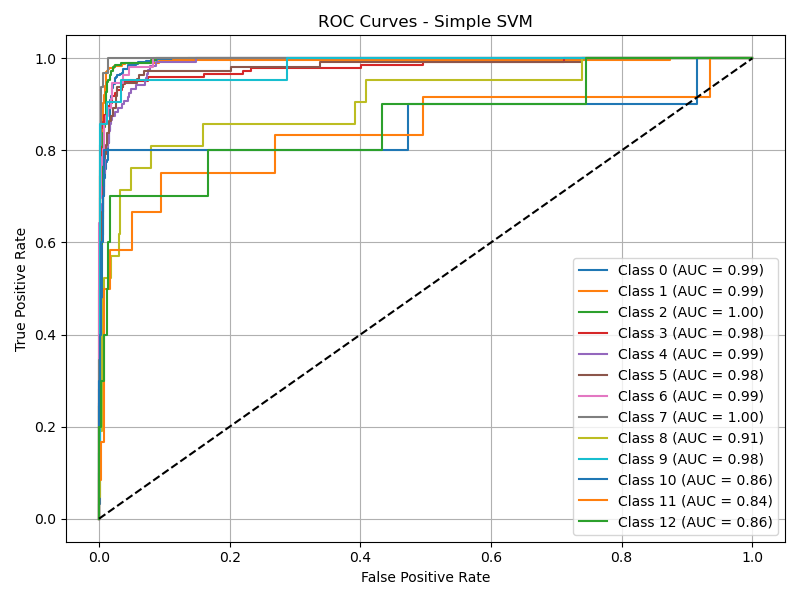}
		\caption{SSVM}
	\end{subfigure}
	\hfill
	\begin{subfigure}[b]{0.24\textwidth}
		\centering
		\includegraphics[width=\textwidth]{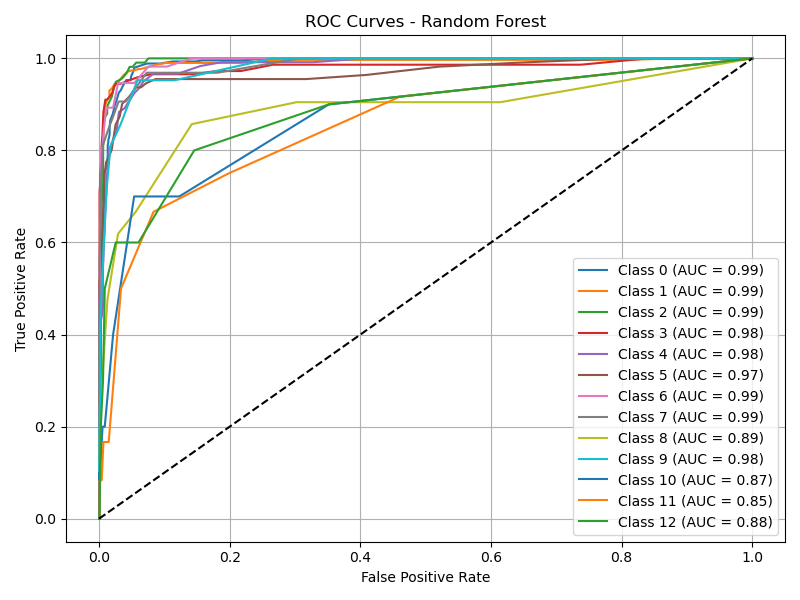}
		\caption{RF}
	\end{subfigure}
	\hfill
	\begin{subfigure}[b]{0.24\textwidth}
		\centering
		\includegraphics[width=\textwidth]{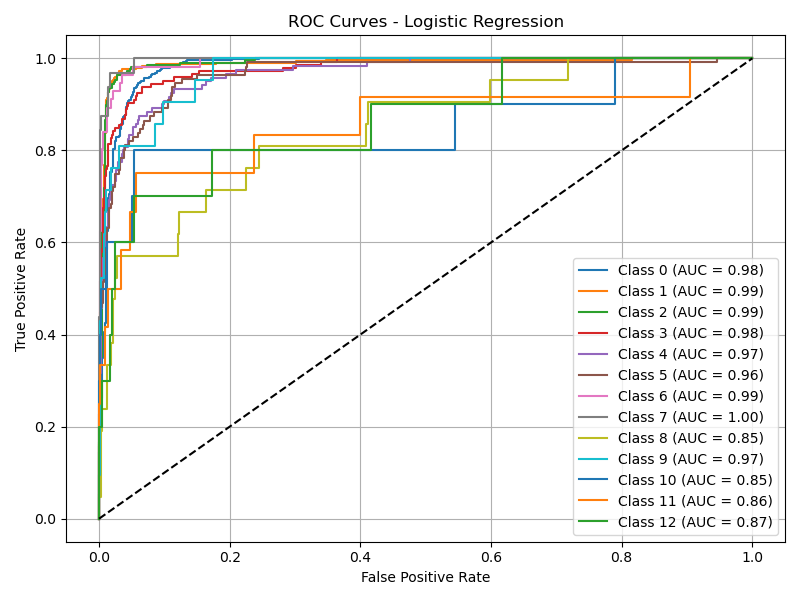}
		\caption{LR}
	\end{subfigure}
	
	\vspace{0.5em}
	
	\begin{subfigure}[b]{0.24\textwidth}
		\centering
		\includegraphics[width=\textwidth]{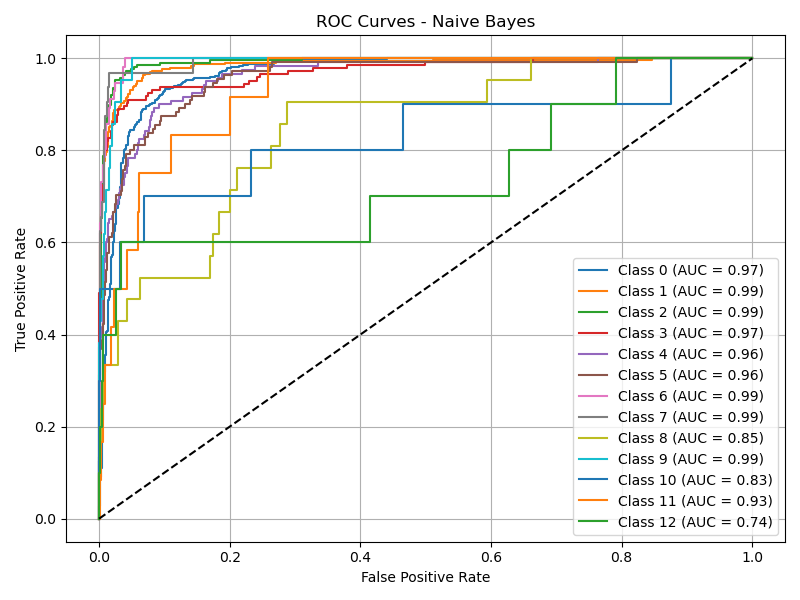}
		\caption{NB}
	\end{subfigure}
	\hfill
	\begin{subfigure}[b]{0.24\textwidth}
		\centering
		\includegraphics[width=\textwidth]{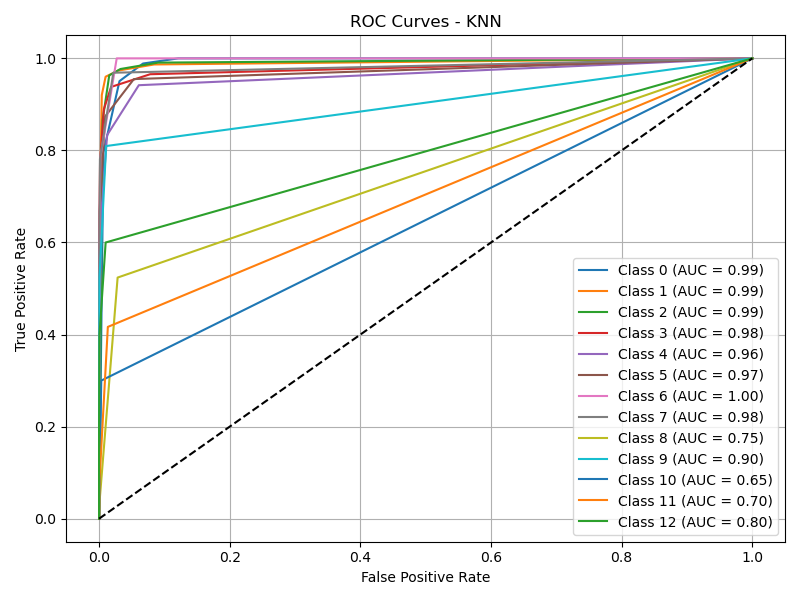}
		\caption{KNN}
	\end{subfigure}
	\hfill
	\begin{subfigure}[b]{0.24\textwidth}
		\centering
		\includegraphics[width=\textwidth]{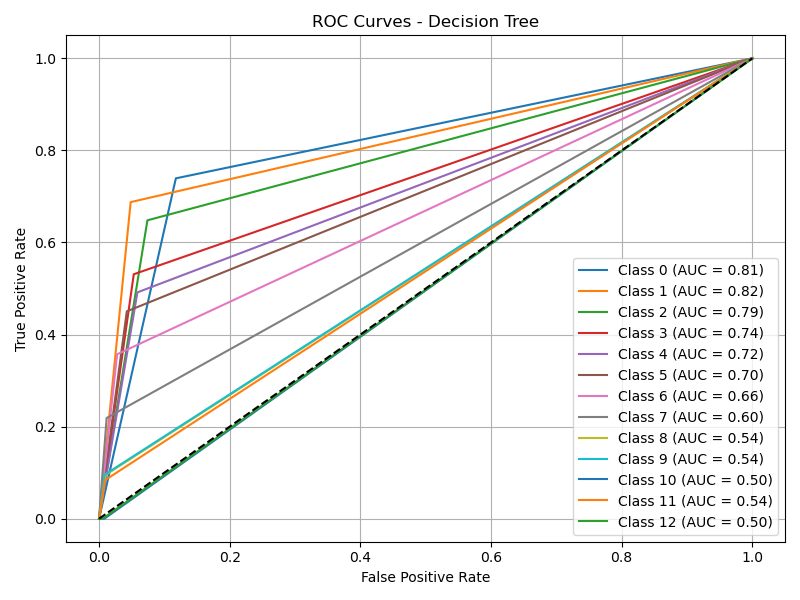}
		\caption{DT}
	\end{subfigure}
	\hfill
	\begin{subfigure}[b]{0.24\textwidth}
		\centering
		\includegraphics[width=\textwidth]{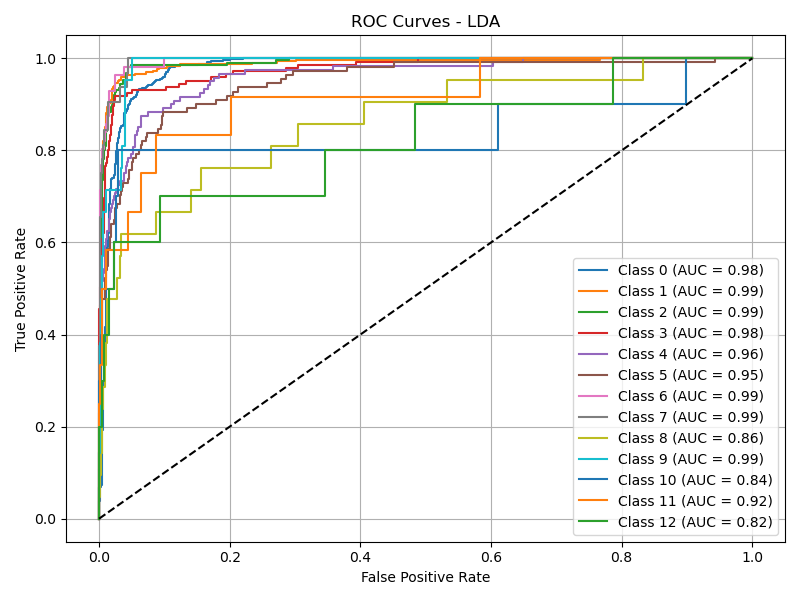}
		\caption{LDA}
	\end{subfigure}
	
	\caption{ROC curves of the proposed classifier alongside those of other benchmark classifiers.}
	\label{fig:roc_curves_all}
\end{figure}

 HPOSVM achieves the highest ROC AUC, indicating superior classification performance. SSVM, LR, and RF also demonstrate strong results, whereas DT and NB exhibit comparatively lower performance.

As shown in TABLE~\ref{tab:performance70}, the proposed HPOSVM model outperformed all other classifiers across all evaluation metrics. It achieved the highest accuracy (0.9327), precision (0.9357), recall (0.9327), F1 score (0.9196), and ROC AUC (0.9618), demonstrating its robustness and reliability for disease detection.

Among the baseline classifiers, SSVM and LR demonstrated strong performance with accuracies of 0.894 and 0.8553, respectively. Ensemble methods such as RF and LDA performed moderately well, while KNN and NB achieved lower but still competitive scores. DT showed the weakest performance, with an accuracy of 0.596 and a ROC AUC of 0.6509.

\section{Conclusion and Future Work}
\label{sec:con}
The proposed CPS demonstrates a reliable and effective approach for monitoring the behavior and physiological conditions of dairy cows in real time. By leveraging a predictive ML model, the system enables early detection of a range of common bovine diseases, including Bloat, BRD, Displaced Abomasum, FMD, Hardware Disease, Johne’s Disease, Ketosis, Lameness, Mastitis, Milk Fever, Tuberculosis, and Acidosis. The CPS integrates multimodal sensor devices, cloud-based data processing, and a user-accessible interface, providing a comprehensive platform for individualized health assessment and continuous disease monitoring in dairy farming environments.

Future research can focus on identifying additional measurable physiological and behavioral indicators associated with a wider range of bovine diseases. This would support the integration of new sensor technologies capable of capturing these indicators, thereby enhancing the diagnostic coverage and predictive accuracy of the proposed framework. Expanding the system in this way would contribute to more comprehensive and precise health monitoring in dairy herds.

	\ifCLASSOPTIONcaptionsoff
	\newpage
	\fi
	
	\bibliographystyle{IEEEtran}
	\bibliography{bibliography.bib}

\end{document}